\documentclass[sigplan,screen]{acmart}
\setcopyright{none}
\renewcommand\footnotetextcopyrightpermission[1]{}

\usepackage{amsmath} 

\usepackage{wrapfig}
\usepackage{booktabs} 
\usepackage{amsmath}
\usepackage{amsfonts}
\usepackage[linesnumbered,commentsnumbered,ruled,vlined]{algorithm2e}

\SetCommentSty{mycommfont}

\SetKwRepeat{Do}{do}{while}

\usepackage{microtype}
\usepackage{multicol}
\usepackage{latexsym}
\usepackage{multirow}
\usepackage{graphicx}
\usepackage{threeparttable}
\usepackage{tikz}
\usetikzlibrary{arrows, automata, positioning}
\usepackage{pgfplots}
\pgfplotsset{compat=1.3}
\usepackage{subfigure}

\usepackage{url}
\usepackage{hyperref}
\usepackage{pifont}
\usepackage{tcolorbox}

\usepackage[utf8]{inputenc}
\usepackage{cleveref}
\crefname{section}{§}{§§}
\Crefname{section}{§}{§§}

\newcommand{\ToolName}{\text{Owl}}

\newcommand{\Number}{1024}

\newcommand{\mybox}[1]{
\begin{tcolorbox}[
boxsep=-0.5pt,
standard jigsaw,
boxrule=0.6pt,
opacityback=0,
sharp corners]
#1
\end{tcolorbox}
}

\usepackage{balance}
\balance
\begin{document}
	
\title{Learning Splitting Heuristics  for Parallel String Solvers}

\author{Chenhao Gao}
\affiliation{%
  \institution{The State Key Laboratory of Blockchain and Data Security, Zhejiang University}
  \city{Hangzhou}
  \country{China}
}
\email{gaochenhao@zju.edu.cn}

\author{Peisen Yao}
\affiliation{%
  \institution{The State Key Laboratory of Blockchain and Data Security, Zhejiang University}
  \city{Hangzhou}
  \country{China}
}
\email{pyaoaa@zju.edu.cn}
\begin{abstract}
String constraint solvers are crucial for reasoning about string-manipulating programs. However, many practical string constraints are undecidable, and real-world applications often present complex constraints that challenge current solvers. The rise of multi-core architectures offers an opportunity for parallel solving. A key parallel solving method is \emph{cube-and-conquer}, in which the quality of splitting heuristics is critical to effectively dividing the search space. Unfortunately, manually designing the heuristics is labor-intensive, and handcrafted heuristics are often sub-optimal.
This paper introduces a data-driven approach to automatically generating splitting heuristics. We frame the problem of selecting a splitting atom as a learning task, using features from input formulas and dynamic data from solver execution. We implement this approach in two popular string solvers, Z3seq and Z3str4, demonstrating that the learned heuristics outperform manually designed ones in the number of solved formulas and the average solving time.
\end{abstract}
\maketitle


\section{Introduction}
\label{sec:introduction}

String manipulation is fundamental to modern software systems, where strings are used extensively for data representation, communication, and user interaction. Ensuring the correctness of string operations is critical, as improper handling can lead to security vulnerabilities and system failures~\cite{eghbali2020no}. Consequently, reasoning about strings has become essential in domains such as cross-site scripting detection~\cite{saxena2010symbolic,trinh2014s3,yu2010stranger}, security policy enforcement~\cite{backes2018semantic}, and validation of data-intensive applications~\cite{gulzar2019white}. To address these needs, a range of SMT solvers for string constraints has been developed~\cite{abdulla2015norn,trinh2014s3,z3seq,zheng2013z3,zheng2017z3str2,Z3str3,aydin2015automata,aziz2017flatten,abdulla2019chain,holik2017string,chen2019decision,liang2014dpll}, supporting operations such as equality, concatenation, length, regular expressions, and replacement. These solvers have enabled advances in symbolic execution~\cite{pyconbyte,li2014symjs} and taint analysis~\cite{zheng2013z3}, among other applications.

Existing work on string constraint solving can be broadly categorized into bounded methods (e.g., fixed-length encodings) and unbounded methods (e.g., automata-based and CDCL($T$)-based solvers). The most widely used solvers for unbounded strings, including CVC4~\cite{barrett2011cvc4}, Z3seq~\cite{z3seq}, Trau~\cite{aziz2017flatten,abdulla2019chain},  Z3str~\cite{zheng2013z3}, Z3str2~\cite{zheng2017z3str2},  Z3str3~\cite{Z3Str33}, are built on the CDCL($T$) framework and support rich string theories. They combine conflict-driven clause learning with theory-specific reasoning for strings and integers, and have become the backbone of many program analysis and verification tools. However, many practical fragments of string constraints remain undecidable~\cite{chen2019decision}, and existing solvers often exhibit limited scalability, thereby posing challenges to their widespread adoption.

The proliferation of many-core architectures has spurred significant interest in parallel constraint solving, primarily through two paradigms: portfolio and divide-and-conquer. The portfolio approach executes multiple solvers concurrently on the same input, returning the result from the first to complete~\cite{marescotti2018smts}. 
Although straightforward to implement, its performance is fundamentally limited by the fastest sequential solver. 
In contrast, the divide-and-conquer paradigm partitions the input formula into subproblems that are solved in parallel, enabling potential speedups beyond sequential baselines~\cite{DBLP:conf/sat/HyvarinenMS15,paropensmt,DBLP:conf/atva/MarescottiHS16}. However, the effectiveness depends critically on the partitioning strategy, as poor decompositions can lead to redundant work or unbalanced workloads.

A prominent, state-of-the-art partitioning technique is \emph{cube-and-conquer}~\cite{heule2016solving,heule2018cube,DBLP:conf/sat/HyvarinenMS15,paropensmt,DBLP:conf/atva/MarescottiHS16,reisenberger2014pboolector}. Given a formula $\varphi$, which is a Boolean combination of a set $S$ of atoms, the cube-and-conquer solver applies a \emph{splitting heuristic} to select an atom $p \in S$. This produces two subformulas, $\varphi \land p$ and $\varphi \land \neg p$, which are solved in parallel. The original formula is satisfiable if at least one subformula is satisfiable. This process can be recursively applied to further increase parallelism.
Existing cube-and-conquer SMT solvers primarily target linear arithmetic and bit-vector arithmetic; Z3~\cite{de2008z3} is the only solver supporting parallel string constraint solving.
Moreover, current splitting heuristics~\cite{paropensmt,DBLP:conf/atva/MarescottiHS16,reisenberger2014pboolector} are typically manually designed, which can struggle with complex, real-world constraints and miss instance-specific optimization opportunities.

This work introduces \ToolName, a data-driven approach for automatically generating effective splitting heuristics in parallel string constraint solving. We recast the problem of selecting an optimal splitting atom as a learning task and address three key challenges faced by prior binary classification-based methods~\cite{DBLP:conf/cp/NejatiFG20}.
First, reducing outcomes to binary labels discards valuable quantitative information, potentially biasing decision-making.
Second, binary classification suffers from cascading errors, where early misclassifications propagate and degrade subsequent decisions, particularly when the number of candidate atoms is large.
Third, designing effective problem features is non-trivial, as they must balance expressiveness and computational efficiency.

To overcome these challenges, \ToolName\ formulates atom selection as a regression problem, directly predicting solving times rather than relying on pairwise classification. This approach preserves the continuity of performance metrics and mitigates error accumulation; moreover, it avoids the linear scan over candidate pairs required by classifier-based ranking. We also design a comprehensive feature set that captures static structural properties (e.g., operator distribution) and dynamic solver states (e.g., conflict statistics). These features enable accurate prediction while maintaining computational efficiency, allowing our approach to adapt to the specific characteristics of each constraint instance.

We have implemented \ToolName\ to accelerate two state-of-the-art string solvers, Z3seq~\cite{z3seq} and Z3str4~\cite{mora2021z3str4}, and evaluated it using diverse benchmark suites of string constraints generated by industrial and academic tools~\cite{saxena2010symbolic,reynolds2017scaling,barrett2010satisfiability}. We train solver-specific regressors and compare against the default lookahead-based splitting heuristics in both backends.
Compared with existing manually crafted splitting heuristics, our learned heuristics solve up to 46 and 9 more formulas for Z3seq and Z3str4, respectively (with 4 threads), and achieve average speedups of 1.44$\times$ and 1.59$\times$ at 32 threads. 
In summary, we make the following key contributions:
\begin{itemize}
    \item We introduce a novel formulation that recasts the design of effective splitting heuristics for parallel string solving as a regression problem.
    \item We propose a practical learning framework that automatically generates splitting heuristics using a rich set of dedicated instance features.
    \item We implement our approach for two string solvers (Z3seq and Z3str4) and conduct a systematic evaluation. Our tool is available at \url{https://tinyurl.com/4yepw9fw}.
\end{itemize}

\section{Preliminaries}
This section presents the basic terminologies used throughout the paper.

\subsection{CDCL($T$) for String Constraint Solving}
\label{subsec:dpllt}

String constraints, a specialized subclass of SMT, have gained attention due to their importance in analyzing string manipulating programs~\cite{saxena2010symbolic, thome2017search}. These solvers are applied in various domains, such as dynamic symbolic execution for dynamically-typed languages~\cite{pyconbyte, li2014symjs}, static taint analysis for Java web applications~\cite{zheng2013z3}, and the verification of access control policies~\cite{rungta2022billion}. Since February 2020, the SMT-LIB standard has incorporated Unicode-based string manipulation and regular expression operations.
Most modern string solvers, including CVC4~\cite{liang2014dpll}, Z3seq~\cite{z3seq}, and Z3str4~\cite{mora2021z3str4}, are built upon the CDCL($T$) framework for SMT solving~\cite{10.5555/647771.734410,sebastiani2007lazy,bruttomesso2007lazy}.

Algorithm~\ref{alg:dpllt} presents the overall procedure of CDCL($T$).
The core idea is to determine an atom's truth value, propagate its consequences, and check for conflicts. If a conflict is found, the algorithm backtracks and learns from it to avoid similar mistakes.
The algorithm is similar to the CDCL algorithm, with the main differences in the functions $decide()$, ${propagate}\_and\_check()$, and $resolve\_conflict()$. 
The function
$decide()$ selects the next unassigned atom and guesses
its value (Line~\ref{line:decide}).
If no atom can be selected, the formula is satisfiable (Line~\ref{line:sat}).
Given the current assignment,
the function $propagate\_and\_check()$ 
applies Boolean constraint propagation and theory propagation~\cite{nieuwenhuis2005dpll} (Line~\ref{line:prop}), which infer values for as many literals and variables as possible.
The function returns false if it encounters a conflict 
and true otherwise. 
In case of a conflict, $resolve\_conflict()$ 
 learns the conflict clauses and performs the backtracking.

\begin{algorithm}[t]
	\DontPrintSemicolon
	\caption{CDCL($T$) SMT solving algorithm.}
	\label{alg:dpllt}
	\KwIn{A first-order formula  $\varphi$}
	\KwOut{SAT or UNSAT}
	\SetKwFunction{lookahead}{look\_ahead}
	\SetKwProg{Fn}{Function}{:}{}
	
	\While{resource limit not reached} {
	    \If{$!decide()$} {
	    \label{line:decide}
            \tcp{Perform branching and phase selection}
			\Return SAT\;
			\label{line:sat}
	    }
	   \While{$!propagate\_and\_check()$} {
	   	\label{line:prop}
                 \tcp{Perform Boolean and theory propagation and backtracking}
	   	  \label{line:conflict0}
	   	  \If{$decision\_level == 0$} { 
	   	     \Return UNSAT\;
    	      } \Else{
           $resolve\_conflict()$\;
                 }
	  }
      }
\end{algorithm}

\begin{example}
\label{exmp:str_alt}
Consider 
$\varphi \equiv x = \texttt{str.++}(\text{``}abc\text{''}, \text{``}d\text{''}) \land (y = \text{``}abcd\text{''} \lor y = \text{``}efg\text{''}) \land x = y$, where $x$ and $y$ are string variables.
Suppose the atom $y = \text{``}efg\text{''}$ is selected by the function $decide()$ and assigned true. The function $propagate\_and\_check()$ can then deduce $x = \text{``}efg\text{''}$ based on the equality $x = y$. However, this assignment is inconsistent with the constraint $x = \texttt{str.++}(\text{``}abc\text{''}, \text{``}d\text{''})$, resulting in a conflict. 
\end{example}

\subsection{Cube-and-Conquer Parallel Solving}
\label{subsec:cube}

A common approach to parallel constraint solving is the divide-and-conquer~\cite{hamadi2018handbook}. Given a formula $\varphi$, the goal is to partition it into a set of subformulas $\psi_1, \ldots, \psi_n$ such that $\varphi \equiv \psi_1 \lor \cdots \lor \psi_n$. If any subformula $\psi_i$ is satisfiable, then the original formula $\varphi$ is satisfiable. If all subformulas are unsatisfiable, then $\varphi$ is unsatisfiable. Each subformula is processed by a sequential solver.


\smallskip
\noindent \textbf{Cube-and-Conquer}.
The partitioning strategy is crucial to the performance of divide-and-conquer solvers~\cite{plaza2008low,hyvarinen2006distribution}. 
A widely used and effective method is the cube-and-conquer strategy~\cite{heule2018cube,DBLP:conf/sat/HyvarinenMS15,paropensmt,DBLP:conf/atva/MarescottiHS16}. In this approach, the search space is divided using a set of cubes—conjunctions of literals—to split the original formula into subformulas.

\begin{definition} [Cube]
	Given a first-order formula $\varphi$, 
	a \textit{cube} $C$ is the conjunction of a set of literals
	$l_1 \land \cdots \land l_n$. 
\end{definition}

\begin{example}
\label{exmp:cube0}
Consider the string formula $\varphi \equiv p \land (q \lor \neg s) \land (r \lor s) \land \cdots \land p$ with four atoms $\{ p, q, r, s \}$.
A cube could be  $p \land q$, $p \land \neg q \land r$, and others.
Intuitively, a cube represents a portion of the search space. For instance, the cube $p \land \neg q$ represents the assignment $p = \text{true}, q = \text{false}$.
\end{example}

At a high level, the cube-and-conquer strategy divides a complex formula into simpler subproblems that can be solved independently in parallel. This division happens by creating a set of cubes $C_1, \ldots, C_n$, where each cube represents a partial assignment of variables. The original formula $\varphi$ is then broken down into subformulas $\varphi \land C_1, \ldots, \varphi \land C_n$, which are distributed to worker solvers.

The key to effective decomposition lies in the \textit{splitting heuristic}, which dynamically decides which atom (Boolean variable or constraint) to branch on at each step of the recursive partitioning process. A good splitting heuristic creates well-balanced subproblems that maximize solving efficiency across all workers.

\begin{figure}[t]
    \centering
    \includegraphics[width=0.48\textwidth]{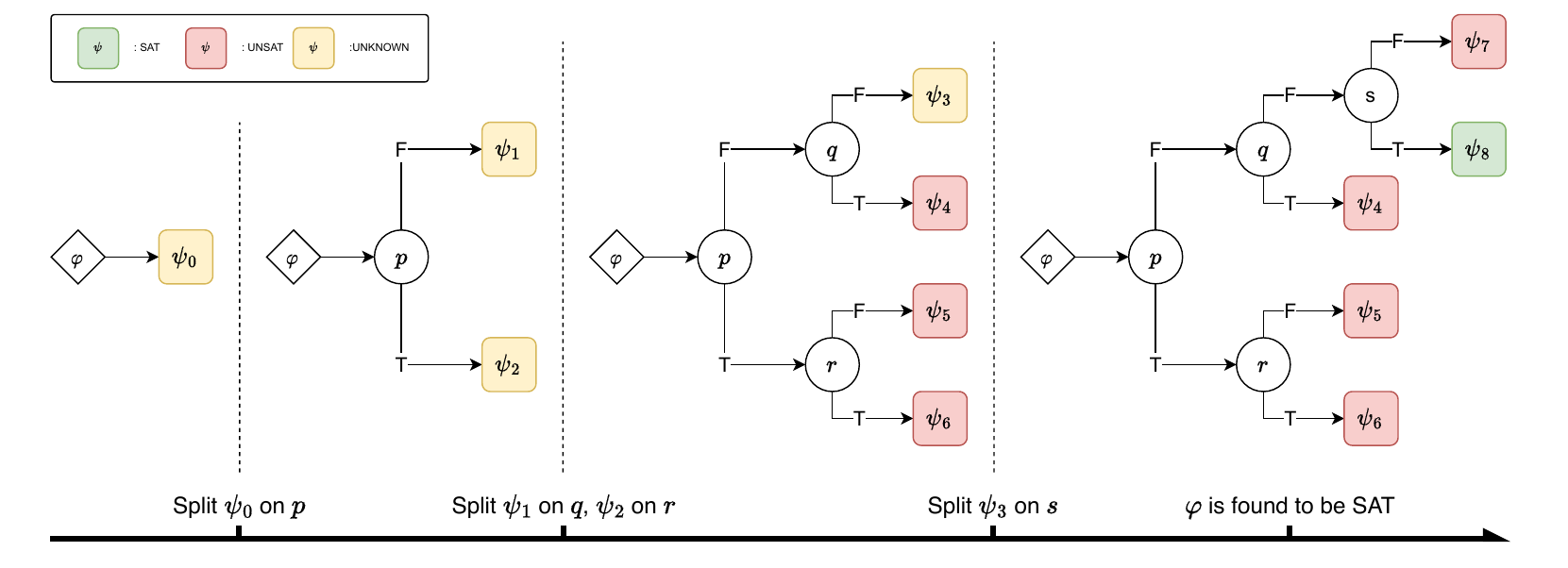}
    \caption{A demonstration of cube-and-conquer}
     \label{fig:cube-and-conquer}
\end{figure}

\begin{example}	 Consider the formula $\varphi$ from Example~\ref{exmp:cube0}. Assume that the atom $p$ is selected for splitting. The solver generates two subformulas: $\psi_1 \equiv \varphi \land p$ and $\psi_2 \equiv \varphi \land \neg p$, which can be solved in parallel. Further splits can be made recursively, for example, by splitting $\psi_1$ on atom $q$, yielding $\psi_3 \equiv \varphi \land p \land q$ and $\psi_4 \equiv \varphi \land p \land \neg q$. Figure~\ref{fig:cube-and-conquer} shows a possible procedure of the cube-and-conquer algorithm. We can observe that $\varphi$ is satisfiable as the subformula $\psi_8$ is satisfiable. \end{example}

\noindent \textbf{Lookahead-based Splitting Heuristic}.
Selecting a splitting atom is critical for the performance of cube-and-conquer solvers~\cite{paropensmt}. However, identifying the optimal splitting atom is challenging and typically addressed through various heuristics. One widely used approach is the \textit{lookahead} heuristic~\cite{paropensmt,DBLP:journals/jsat/HeuleM06}, which estimates the solver's behavior after splitting on different candidate atoms.

Algorithm~\ref{alg:lookahead} is a high-level description of the lookahead-based splitting heuristic.
Given a formula $\varphi$ and a set $A$ of candidate atoms,
the solver attempts to split on each atom $p_i \in A$ and perform the constraint propagation for limited steps (Lines~\ref{line:prop0}-\ref{line:prop1}).
Based on the information gathered during propagation (e.g., the number of learned clauses), the algorithm computes a non-negative score $score_i$ for each atom $p_i$ (Line~\ref{line:score}). The score reflects the potential impact of splitting on $p_i$. The atom with the highest score is then selected as the splitting atom.

\begin{algorithm}[t]
	\DontPrintSemicolon
	\caption{Lookahead-based splitting heuristic.}
	\label{alg:lookahead}
	\KwIn{A formula  $\varphi$ and a set of atoms $A =  \{p_1, \ldots, p_n \}$}
	\KwOut{An atom in  $A$ for the solver to split on}
	\SetKwFunction{lookahead}{look\_ahead}
	\SetKwProg{Fn}{Function}{:}{}
     $all\_scores \leftarrow []$\; 
	\For{$i = 1 $ to $n$}{	
			Propagate $\varphi \land p_i$ for limited steps\;
			\label{line:prop0}
			Propagate $\varphi \land \neg p_i$ for limited steps\;
			\label{line:prop1}

			${score}_i \leftarrow$ Compute a score for $p_i$ using the information during the constraint propagation\;   			
            \label{line:score}
             $all\_scores$.append(${score}_i$)
	}
        $best\_idx = arg max(all\_scores)$\;
	\Return \ $p_{best\_idx}$\;

\end{algorithm}

\section{Overview}
\label{sec:motivation}
This section formulates the problem of optimizing the splitting heuristic (\cref{subsec:probstatement}), discusses the limitations of existing work (\cref{subsec:existing}), and outlines our solution (\cref{subsec:ours}).

\subsection{Splitting Heuristic Optimization}
\label{subsec:probstatement}
Designing effective splitting heuristics for SMT solvers is challenging because it requires deep domain knowledge and extensive manual tuning. 
Our goal is to automate this process and develop data-driven heuristics by adapting to the specific characteristics of each problem instance.
To formalize it, let $\varphi$ be a formula and $S$ be a cube-and-conquer parallel SMT solver that performs a sequence of splits on atoms.
We define $Cost(\varphi, \mathcal{P})$ as the oracle that reports the cost for $S$ to solve $\varphi$ given a sequence of splits on atoms $\mathcal{P} = (p_1, p_2, \ldots, p_n)$.
Then, the objective is to select a sequence of atoms that minimizes this cost:
\begin{equation}
\label{eqn:prob}
\operatorname*{arg\,min}_{\mathcal{P} \in \mathcal{A}} Cost(\varphi, \mathcal{P}),
\end{equation}
where $\mathcal{A} \subseteq \text{Atoms}(\varphi)^*$ represents the space of possible atom sequences that can be selected during the solving process.

However, finding the optimal strategy is computationally infeasible. Even for a formula with 100 atoms, the number of possible branching sequences grows combinatorially with the branching depth. Moreover, evaluating $Cost(\varphi, \mathcal{P})$ for each candidate sequence would require running the solver on an exponential number of subformulas, which is prohibitively expensive.
In this work, we adopt a more tractable greedy selection approach. At each decision point along a given branch, we select the next splitting atom $p$ that is expected to minimize the remaining solving cost:
\begin{equation}
\operatorname*{arg\,min}_{p \in \text{Atoms}(\varphi)} \,\mathbb{E}\Bigl[Cost\Bigl(\varphi, (\mathcal{P}_{i-1}, p)\Bigr)\Bigr],
\end{equation}
where $\mathcal{P}_{i-1}$ represents the sequence of atoms that have been selected so far along that branch, and $(\mathcal{P}_{i-1}, p)$ denotes sequence extension by appending $p$.
While this approach does not yield a globally optimal branching strategy, it enables iterative refinement of splitting decisions at the level of individual subformulas. 

\subsection{Limitations of Existing Work}
\label{subsec:existing}
Existing cube-and-conquer-based SMT solvers~\cite{paropensmt,DBLP:conf/atva/MarescottiHS16,reisenberger2014pboolector} employ manually crafted heuristics for selecting splitting atoms. These heuristics are typically static and specific to the heuristic, making limited use of instance-specific structure.

In the context of parallel SAT solving, Nejati et al.~\cite{DBLP:conf/cp/NejatiFG20} propose an automated approach based on pairwise classification. Their model is a binary classifier that compares two candidate atoms $(X_i, X_j)$ at a time, using associated solving times $(t_i, t_j)$, to identify the more advantageous branching choice. However, this formulation requires repeatedly invoking the classifier over candidate pairs, resulting in linear (or worse) scaling with the number of atoms, similar to traditional lookahead-based ranking strategies.

Though the above strategy represents an improvement over fixed heuristics, it has several shortcomings:
\begin{itemize}
    \item 
    \emph{Loss of Quantitative Information}: 
    Reducing real-valued solving times to binary preferences discards fine-grained cost information. This quantization limits the model's expressivity and impairs its ability to discriminate among closely ranked candidates.
    \item 
    \emph{Accumulated Error}: 
     The use of serial comparisons introduces compounding error. Misrankings at early stages can propagate, particularly when applied to large candidate sets, degrading overall prediction quality.
    
\end{itemize}

Finally, the pairwise approach has been validated only in the SAT domain. Its applicability to SMT solving, especially for theories such as strings, arrays, or mixed arithmetic, remains unproven. In our evaluation (\cref{sec:evaluation}), we include this method as a baseline and observe that its performance degrades significantly in SMT settings.

\subsection{Our Approach}
\label{subsec:ours}

\noindent \textbf{Formulating the Costs}.
Rather than relying on pairwise classification, we model solving time as a continuous quantity. This approach preserves the full numerical range of solution costs and provides more precise cost estimates for each potential splitting atom.
Formally, given a formula $\varphi$ and a candidate splitting atom $p$, let $\psi_1 = \varphi \land p$ and $\psi_2 = \varphi \land \neg p$ be the subformulas resulting from the split. We define the total solving time $Time(\varphi, p)$:

\begin{equation}
	Time(\varphi, p) = \left\{
	\begin{aligned}
		min(t_1, t_2), & \ \ \psi_1: \text{SAT}, \psi_2: \text{SAT} \\
	    t_1, & \ \ \psi_1:  \text{SAT}, \psi_2: \text{UNSAT} \\
            t_2, & \ \ \psi_1: \text{UNSAT}, \psi_2: \text{SAT} \\
            max(t_1, t_2), & \ \ \psi_1: \text{UNSAT}, \psi_2: \text{UNSAT} \\
            sum(t_1, t_2), & \ \ \psi_1: \text{UNSAT}, \psi_2: \text{UNKNOWN} \\
                           & \ \ \psi_1: \text{UNKNOWN}, \psi_2: \text{UNSAT} \\
                           & \ \ \psi_1: \text{UNKNOWN}, \psi_2: \text{UNKNOWN} \\
	\end{aligned}
	\right.
	\label{eqtime}
\end{equation}
\normalsize

This formulation reflects the key insights of parallel solving. In SAT–SAT cases, solving completes upon the first solution; in SAT–UNSAT cases, the satisfiable branch dominates the runtime; in UNSAT–UNSAT cases, both branches must be exhaustively explored; and in the presence of timeouts (UNKNOWN), the total cost accumulates across all solver invocations. Rather than relying on binary comparisons over branching literals, we model solving time directly, preserving richer structural information. This approach better guides branch selection by accounting for the asymmetric and outcome-sensitive complexity of solving two subformulas with three-valued results.

\smallskip
\noindent \textbf{Designing Problem Features}.
To support effective learning, we introduce a 77-feature representation, partitioned into 66 static and 11 dynamic features.
 The static features capture structural properties of $\varphi$.
 The dynamic features reflect real-time solver states (e.g., lookahead and lookback statistics) that update as new information becomes available.
These features encode the formula's underlying structure and the solver's changing states, enabling our learning approach to adjust the splitting heuristic \emph{in situ}.

\smallskip
\noindent \textbf{Learning Splitting Heuristics}.
We formulate splitting-atom selection as a \emph{regression} task: given a formula $\varphi$ and atom $p$, the model predicts $Time(\varphi, p)$, the time required to solve the subproblems induced by branching on $p$. This formulation captures the continuous nature of solver runtimes, avoiding the discretization and relative-error accumulation inherent in pairwise classification. Moreover, by eliminating the need for exhaustive comparisons among candidate atoms, the approach reduces inference overhead, thereby improving both predictive accuracy and runtime scalability.

\section{Methodology}
\label{sec:design}

This section presents our formulation of splitting heuristic optimization as a regression-based problem (\cref{subsec:predict}), details the feature engineering process (\cref{subsec:features}), and describes the model selection and training methodology (\cref{subsec:train}).

\subsection{Regressor-Based Selection}
\label{subsec:predict}

A key limitation of the previous classifier-based approach (cf. \cref{subsec:existing}) is its reliance on pairwise classification schemes, which can obscure valuable information embedded in continuous solving times. Additionally, classification-based methods introduce dependencies between decisions, leading to accumulated errors. To address these issues, we propose a regressor-based method that learns a continuous surrogate of the solver’s behavior.

\smallskip
\noindent \textbf{Problem Formulation}. 
Formally, let $X_i$ denote the feature vector (detailed in \cref{subsec:features}) extracted for the $i$-th atom in a given formula, and let $t_i$ be its observed solving time. 
We gather training examples $\bigl\{(X_i,\,t_i)\bigr\}_{i=1}^N$ from multiple formula instances and split assignments in a data collection phase.
Under a parameterized model $\mathcal{T}_{\theta}$, the learning objective is to minimize the mean squared error (MSE) between predicted and observed times. In other words, regression here means learning a function that maps each feature vector to a numeric runtime estimate:
\begin{equation}
    \theta^* 
    = \operatorname{argmin}_{\theta}
    \frac{1}{N} \sum_{i=1}^N \Bigl(\mathcal{T}_{\theta}(X_i) - t_i \Bigr)^2,
\end{equation}
where $\mathcal{T}_{\theta}(X_i)$ denotes the predicted solving time for atom $X_i$. 
By directly predicting solving times rather than classifying atoms into discrete categories, we retain fine-grained performance information and eliminate interdependencies that can degrade accuracy~\cite{DBLP:conf/cp/NejatiFG20}.

\begin{algorithm}[t]
\DontPrintSemicolon
\caption{A Regressor-Based Splitting Heuristic}
\label{alg:rank-via-vote}
\KwIn{Formula $\varphi$, set of candidate atoms $A = \{p_1, \ldots, p_n\}$, regressor $\mathit{Model}$}%
\KwOut{Index of the predicted best splitting atom in $A$}%
{\sf Features} $\leftarrow [\,]$; \\
\For{$i = 1 \text{ to } n$}{
    Compute lookback features $Feat_{back}$ for $p_i$; \\
    Propagate $\varphi \land p_i$ for limited steps; \\
    Propagate $\varphi \land \neg p_i$ for limited steps; \\
    Compute lookahead features $Feat_{ahead}$ for $p_i$; \\
    {\sf Features}.append($Feat_{back} \,\cup\, Feat_{ahead}$)
}
{\sf Ranks} $\leftarrow \mathit{Model}.{\sf predict}(\textsf{Features})$; \\
\Return $\operatorname{argmin} (\mathsf{Ranks})$;
\end{algorithm}

\smallskip
\noindent \textbf{Plugging into Cube-and-Conquer}.
As shown in Algorithm~\ref{alg:rank-via-vote}, our regressor-based approach first extracts features from each candidate atom and then applies a trained model to estimate its solving time. These estimates are used to produce an \emph{independent} ranking over the candidates.
Modeling solving time as a continuous variable introduces natural separation between candidates: atoms that yield little or no progress—often resulting in an ``UNKNOWN'' outcome—are assigned significantly higher predicted times than those that enable substantial progress. This separation
enhances the reliability of identifying promising atoms and supports more effective pruning in the Cube-and-Conquer framework.

While precisely modeling the full runtime distribution of a modern CDCL($T$)-based SMT solver remains challenging, our pragmatic goal is to distinguish candidate atoms with significantly different solving times.  Our experiments confirm that this approach robustly captures sparse yet critical runtime variations often overlooked by classification-based methods. 

\subsection{Feature Engineering}
\label{subsec:features}
Choosing a suitable feature set $Feat(\varphi, p)$ is crucial: it must be both \emph{predictive} of solving performance and \emph{efficient} to compute. 
As mentioned in \cref{subsec:ours}, we propose a total of 77 features (Table~\ref{str-feat}) that encapsulate both static and dynamic interface points in the solver.

\begin{table}[t]
	\centering
	\caption{Problem features $Feat(\varphi, p)$.}
	\label{str-feat}
	\begin{minipage} {0.47\textwidth}
		\centering
		\textbf{Dynamic Features (1-11)}
		
		\begin{tabular}{c l}
			\toprule
			\textbf{ID} & \textbf{Description} \\
			\midrule		
			1 & \# of times $p$ is assigned \\
			2 & \# of times $p$ is assigned a different \\
			  & value than its cached one \\
			3 & \# of $p$ appears in a conflict 
			   clause \\
			4 & \# of $p$ appears in a lemma \\
			5 & \# of obtained lemma in the 
			   lookahead \\
			6 & \# of assignments in the 
			   lookahead \\
			7 & \# of satisfied clauses in the 
			   lookahead \\
			8 & Lookahead score from Z3 \\
			9 & Average activities of lemmas \\
			10 & \# of propagations \\
			11 & Fraction of \# of conflicts 
			    over \# of decisions \\
			\bottomrule
		\end{tabular}
	\end{minipage}
	\hfill
	\begin{minipage}{0.47\textwidth}
		\centering
		\textbf{Static Features (12-77)}
		
		\begin{tabular}{c l}
			\toprule
			\textbf{ID} & \textbf{Description} \\
			\midrule		
			12 & \# of $p$ appears in a 2-clause \\
			13 & \# of $p$ appears in a 3-clause \\
			14 & \# of $p$ appears in a 4-clause \\
			15 & \# of $p$ appears in the formula \\
			16 & \# of sub-atoms in $p$ \\
			17-45 & One-hot encoding for the
			       atom's type \\
			46-74 & \# of all types of sub-atoms 
			       in $p$ \\
			75 & Average clause degree in the 
			    variable-clause graph \\
			76 & Average variable degree in the 
			    variable-clause graph \\
			77 & \# of atoms in the formula \\
			\bottomrule
		\end{tabular}
	\end{minipage}
\end{table}

\smallskip
\noindent \textbf{Static Features}.
Static features (features 12--77 in Table~\ref{str-feat}) are derived from the formulas and the atom’s structure, such as clause sizes, operator types, or degrees in the variable-clause graph. 
These can be computed without running the solver and offer a baseline indication of complexity. 
For instance, features 46--74 measure occurrence counts of specific operations, which have been recognized as key cost drivers in string-based SMT solvers~\cite{chen2019decision,reynolds2018rewrites}.

\smallskip
\noindent \textbf{Dynamic Features}.
While static properties provide an initial estimate, dynamic features (1--11 in Table~\ref{str-feat}) capture the solver’s current state via short-term \emph{lookahead} and \emph{lookback} heuristics. 
\emph{Lookback heuristics} evaluate an atom’s past performance during the solving process (e.g., how often $p$ has participated in conflict clauses or learned lemmas). 
This heuristic uses the atom’s historical activity to predict its future performance. 
For example, if an atom has been involved in many conflict clauses or learned lemmas, it will likely play a significant role in future solving steps.
\emph{Lookahead heuristics} examine the potential future consequences of splitting on a given atom. We gather the information about how the formula changes in response to these assignments by propagating assignments for a limited number of steps. 
This information can reveal implicit relationships between atoms in the formula, helping to predict how efficient future solving will be.
Similar features have been spotlighted in portfolio-based CSP and SAT solving~\cite{xu2008satzilla,kadioglu2010isac} as strong indicators of future behavior.

This integration of static structural properties with dynamic solver states underpins our regressor's ability to adapt the splitting heuristic during execution.

\begin{example}
\label{ex:propagation}
Consider the Boolean formula $\varphi \equiv p \land (\neg p \lor q) \land (\neg q \lor r)$. 
When we split on $p$ and propagate $\varphi \land p$ (i.e., $p = \text{true}$), additional assignments $q = \text{true}$ and $r = \text{true}$ follow from unit propagation. 
Hence, dynamic features track how many new assignments or conflicts were triggered by setting $p = \text{true}$. 
Similarly, $\varphi \land \neg p$ might reveal a different propagation pattern.
\end{example}

Extracting dynamic features incurs some computational overhead. However, this process relies on bounded propagation with an early-termination threshold (typically around 600 conflicts), rather than on a full CDCL($T$) search. Moreover, the technique aligns with established lookahead-based heuristics used by existing solvers. Empirical results indicate that, despite the added cost, the improved atom selection leads to gains in overall performance.

\subsection{Model Selection and Training}
\label{subsec:train}

\noindent \textbf{Model Selection}.
To enable real-time deployment, our approach prioritizes both \emph{efficiency} and \emph{accuracy}. The model must deliver predictions with minimal latency to avoid negating the performance improvements gained through enhanced splitting decisions.
For this purpose, we adopt Random Forest Regression as the core predictive model in our experiments. Random Forests strike a favorable balance between predictive accuracy and inference speed, making them well-suited for our requirements. Additionally, they exhibit robustness to noisy labels, which is critical given the inherent variability in solving times~\cite{caruana2006empirical, louppe2014understanding}. 
The ensemble-based structure of Random Forests also facilitates feature importance analysis, enabling systematic refinement of the feature set in subsequent iterations when necessary~\cite{caruana2006empirical, louppe2014understanding}.

\begin{figure}[t]
    \centering
    \includegraphics[width=0.49\textwidth]{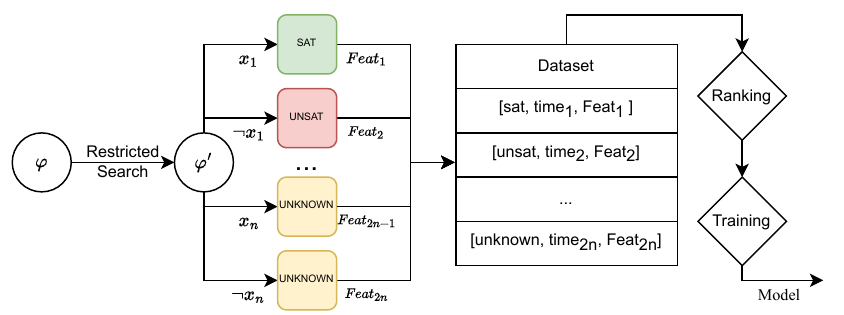}
    \caption{A demonstration of the training set extraction.}
    \label{fig:training}
\end{figure}

\smallskip
\noindent \textbf{Training Set Extraction}.
A representative and diverse training set is essential for effective regression. As shown in Figure~\ref{fig:training}, we collect training data by executing the solver on multiple \emph{distinct} splitting atoms within each formula and measuring the resulting \emph{time-to-solve} for the corresponding subproblems. Repeating this process across a broad benchmark suite yields a dataset of $(X_i, t_i)$ pairs, where $X_i$ encodes structural features of the atom and formula context, and $t_i$ records the post-split solving time. To avoid noise from degenerate cases, we apply Z3’s simplification tactics as a preprocessing filter to eliminate trivially solvable formulas prior to data collection. This ensures that the training distribution emphasizes nontrivial, solver-relevant branching decisions. 

\smallskip
\noindent \textbf{Model Deployment}.
The learned model is integrated into the solver as a component that is periodically invoked to guide branching decisions. At each invocation, the model selects the most promising splitting atom among a set of candidates based on current structural features. To enable efficient use in a parallel setting, we launch $n$ Python worker processes, each dedicated to serving one of the solver’s threads. Solver threads communicate with their worker via inter-process communication (IPC) using bidirectional pipes. When a prediction is required, the solver constructs a feature matrix representing the current candidate atoms and sends it to its designated worker, which performs a batched model evaluation. The worker returns the predicted best atom, which the solver then uses to proceed, following the logic described in Algorithm~\ref{alg:rank-via-vote}. This setup ensures low-latency, parallel predictions while maintaining isolation between solver and model execution.

\section{Evaluation}
\label{sec:evaluation}
We evaluate \ToolName\ on two state-of-the-art string solvers (Z3seq~\cite{z3seq} and Z3str4~\cite{mora2021z3str4}) and investigate the following research questions:
\begin{itemize}
    \item \textbf{RQ1}: How effectively does \ToolName\ improve the two existing parallel string constraint solvers (Z3seq and Z3str4) (\cref{subsec:eval-speedup})
    \item \textbf{RQ2}: How accurately and fast is \ToolName\ in predicting the splitting atoms for Z3seq and Z3str4, and what are the important features? (\cref{subsec:eval-performance})
    \item \textbf{RQ3}: How does the choice of machine learning algorithms affect solver performance? (\cref{subsec:eval-ml})
\end{itemize}

\noindent \textbf{Benchmarks}.  
We evaluate \ToolName\ on 
\Number\ QF\_SLIA formulas drawn from the SMT-LIB repository~\cite{barrett2010satisfiability}, including benchmarks from PyEx, StringFuzz, and PyConbyte. These benchmarks are widely used in prior work~\cite{reynolds2017scaling,DBLP:conf/pldi/AbdullaACDDJLHW20,DBLP:journals/pacmpl/TrinhCJ20,reynolds2018rewrites,reynolds2019high,abdulla2019chain}. 
We exclude constraints that can be solved within 1 second by either Z3seq or Z3str4.
We partition the data into 80/20 stratified splits for training and evaluation. Table~\ref{tbl:stat} summarizes the characteristics of the benchmarks.

\begin{table}[t]
	\caption{Statistics of the string formulas in our evaluation.}
	\label{tbl:stat}
	\centering
	\begin{tabular}{l c c c c }
		\toprule
		\textbf{Statistic}    & \textbf{AVG}. & \textbf{Median}&  \textbf{MAX}. & \textbf{MIN}. \\ 
		\midrule
		\# Atoms  &  651 & 511  &  1684  & 14   \\
		\bottomrule
	\end{tabular}
\end{table}

\smallskip
\noindent \textbf{Methodology}. 
We train a random forest regressor and a random forest classifier using a grid search for parameter tuning (\text{n\_estimators} in [50--300], \text{max\_features}$=$\text{log2}, \text{random\_state}$=$42).
To answer RQ1 and RQ2, we measure solver performance (e.g., execution time, number of solved instances, Par-2 score, and model quality in selecting ``optimal'' splitting atoms. Here, \emph{accuracy} refers to how highly the predicted best atom ranks in the ground-truth runtime ordering, and \emph{efficiency} refers to the time required to compute the prediction. For RQ3, we compare different ML models to quantify this trade-off. 
All experiments are run on an AMD EPYC 1.50 GHz (96 cores) machine with 755 GB of RAM and Ubuntu 22.04, with a per-instance timeout of 120 seconds.

\subsection{Impact on Z3seq and Z3str4 (RQ1)}
\label{subsec:eval-speedup}
\begin{figure*}[t]
	\centering 
	\subfigure[]{\label{fig:Z3seq}%
	\resizebox{0.3\textheight}{!}
		{
			\centering
			\includegraphics[width=\linewidth]{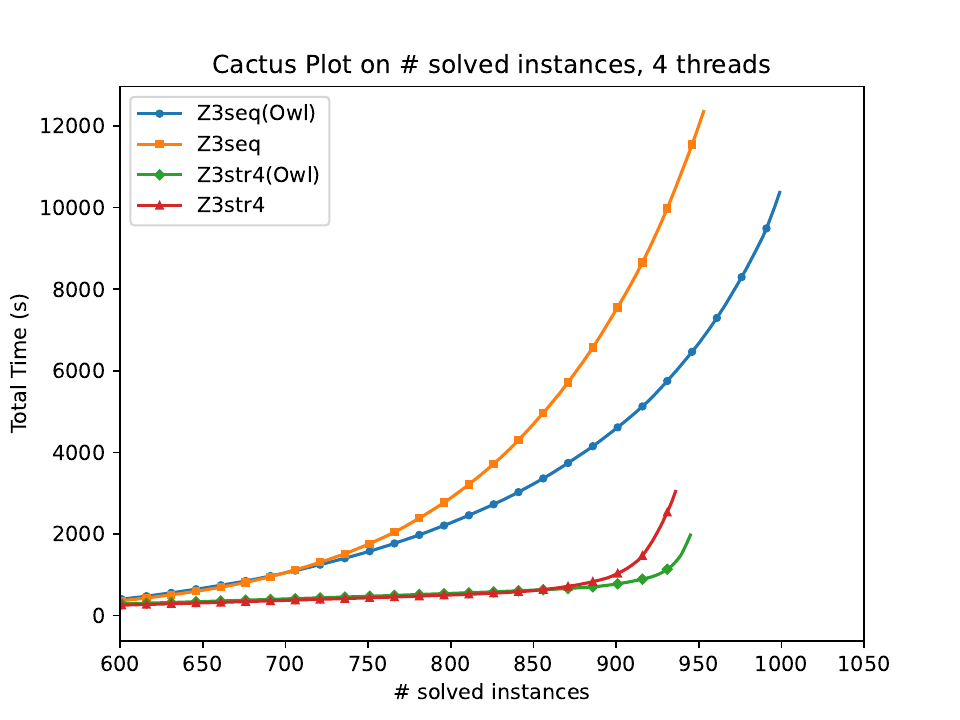}	
		}
	}
	\subfigure[]{\label{fig:Z3str4}%
	\resizebox{0.3\textheight}{!}
		{
			\centering
			\includegraphics[width=\linewidth]{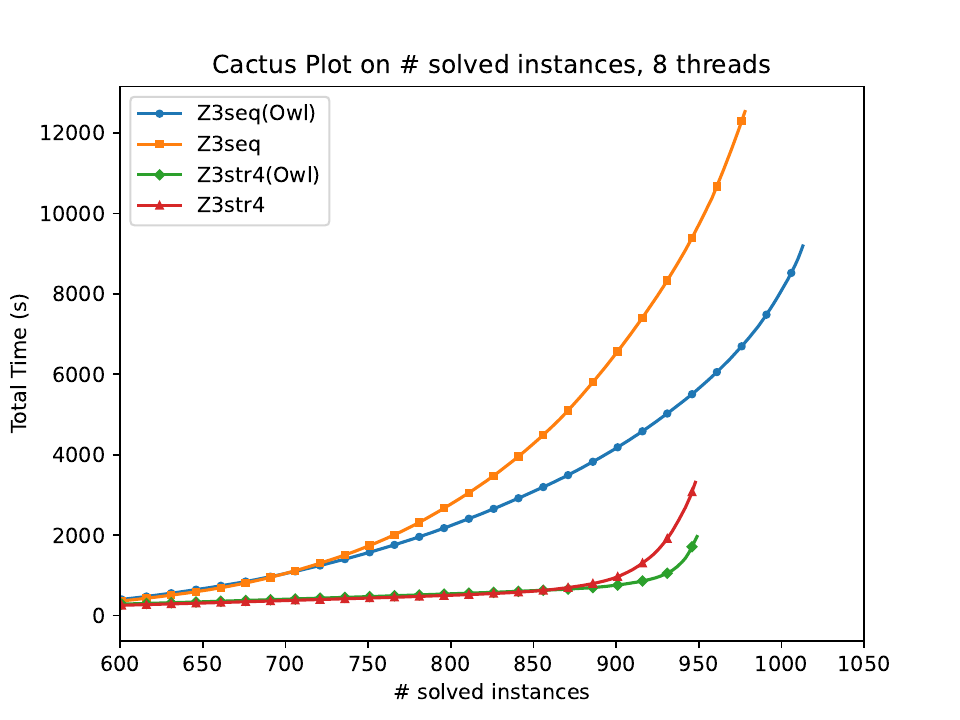}
		}
	}
  	\subfigure[]{\label{fig:Z3str4-16}%
	\resizebox{0.3\textheight}{!}
		{
			\centering
			\includegraphics[width=\linewidth]{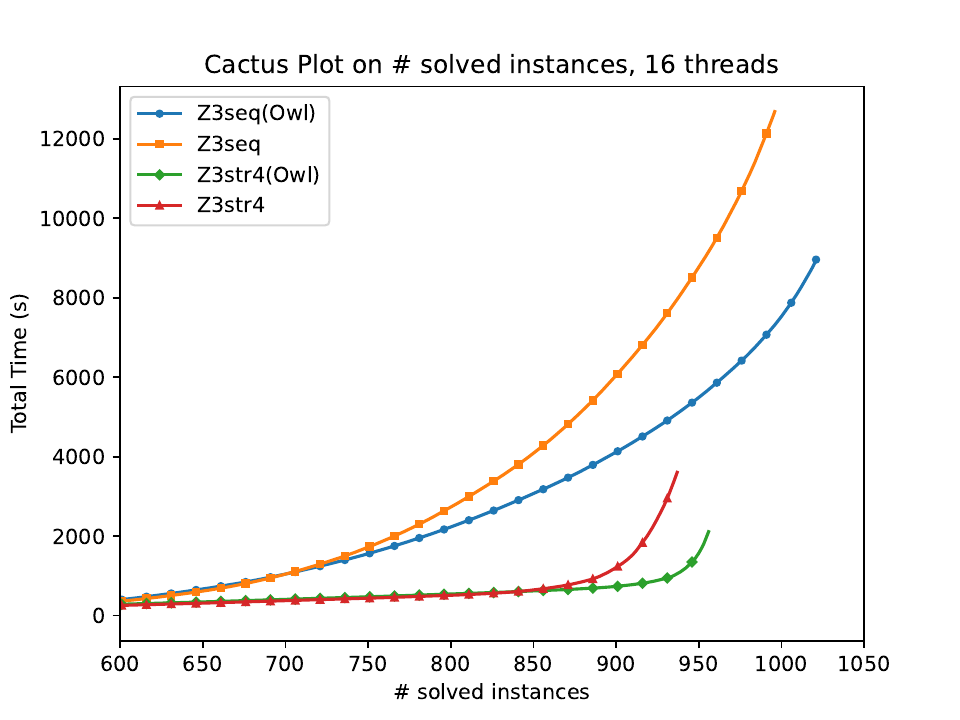}
		}
	}
  	\subfigure[]{\label{fig:Z3str4-32}%
	\resizebox{0.3\textheight}{!}
		{
			\centering
			\includegraphics[width=\linewidth]{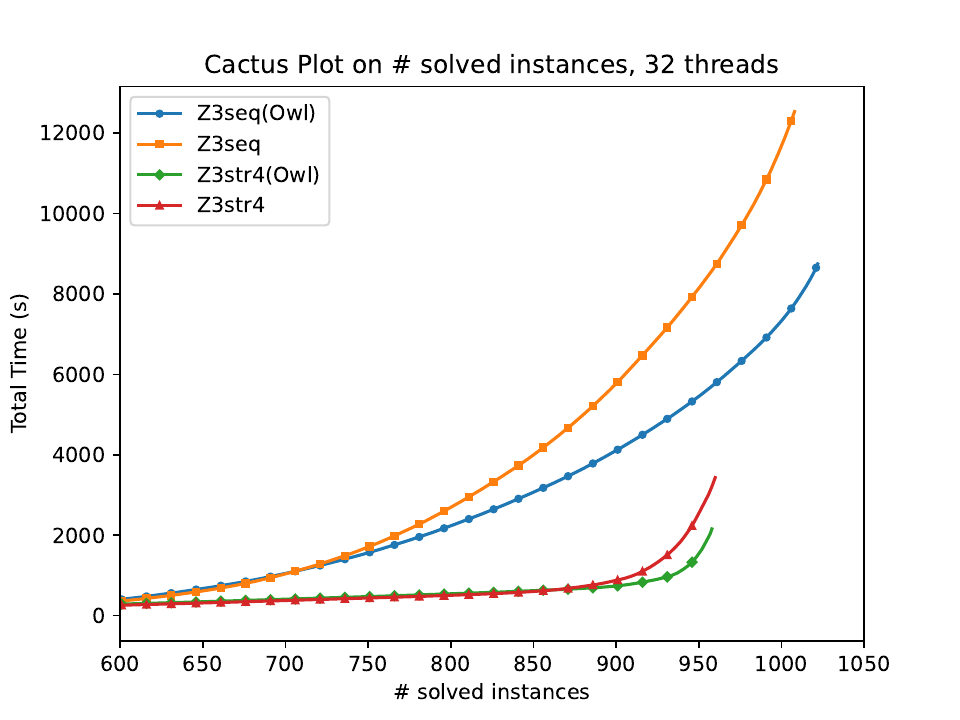}
		}
	}
	\caption{Cactus plot for Z3seq, Z3seq(\ToolName), Z3str4 and Z3str4(\ToolName)}
	\label{fig:cactus}
\end{figure*}

We compare the effectiveness of the learned splitting heuristics with the default strategies in Z3seq and Z3str4.
We use the same solver configurations for a fair comparison, differing only in the splitting heuristics employed and in the number of threads, which we vary from 4 to 32. 
Table~\ref{tbl:seqstr3} summarizes the results: \#~Solved indicates how many formulas are solved before the timeout, Avg.~Time is over all solved instances, and Par-2 Score encodes penalties for unsolved formulas.  Figure~\ref{fig:cactus} provides corresponding cactus plots.

\begin{table}[t]
	\centering
	\caption{Performance with (and without) \ToolName.}
	\label{tbl:seqstr3}

  \resizebox{0.49\textwidth}{!}
   {
	\begin{tabular}{ c  l  c c c }
		\toprule
		\textbf{Threads}            & \textbf{Solver}        & \textbf{\# Solved} & \textbf{Avg. Time} & \textbf{Par-2 Score }\\ 
		\midrule
		\multirow{4}{*}{32} 
		& Z3seq              & 1008     & 12.43   &  16,134 \\ 
		& Z3seq(\ToolName)   & \textbf{1022}     & 8.55    & 9,226 \\ 
		& Z3str4             & 960     & 3.57   &  10,703 \\ 
		& Z3str4(\ToolName)  & 958     & \text{2.25}   &  6,814 \\ 
		\hline
		\multirow{4}{*}{16} 
		& Z3seq               & 996     & 12.74    & 19,165 \\ 
            & Z3seq(\ToolName)    & \textbf{1021}     & 8.77   & 9,680  \\ 
            & Z3str4              & 937      & 3.85   &  11,637 \\ 
            & Z3str4(\ToolName)   & 956      & \text{2.21}   &  7,487 \\ 
            \hline
		\multirow{4}{*}{8} 
		& Z3seq               & 978     & 12.81    & 23,333 \\ 
            & Z3seq(\ToolName)    & \textbf{1013}     & 9.07   & 11,831   \\ 
            & Z3str4              & 948      & 3.50   &  12,274 \\ 
            & Z3str4(\ToolName)   & 949      & \text{2.07}   & 7,594  \\ 
            \hline
		\multirow{4}{*}{4} 
		& Z3seq               & 953     & 12.96    & 29,152 \\ 
            & Z3seq(\ToolName)    & \textbf{999}     & 10.38   & 16,370  \\ 
            & Z3str4              & 936      & 3.25   &  13,931 \\ 
            & Z3str4(\ToolName)   & 945      & \text{2.09}   &  7,374
            \\ 
		\bottomrule
	\end{tabular}
      }
\end{table}

\pgfplotsset{%
 width=.45\textwidth,
  height=0.26\textheight
}

\begin{figure*}[t]
	\centering 
        \includegraphics[width=0.9\linewidth]{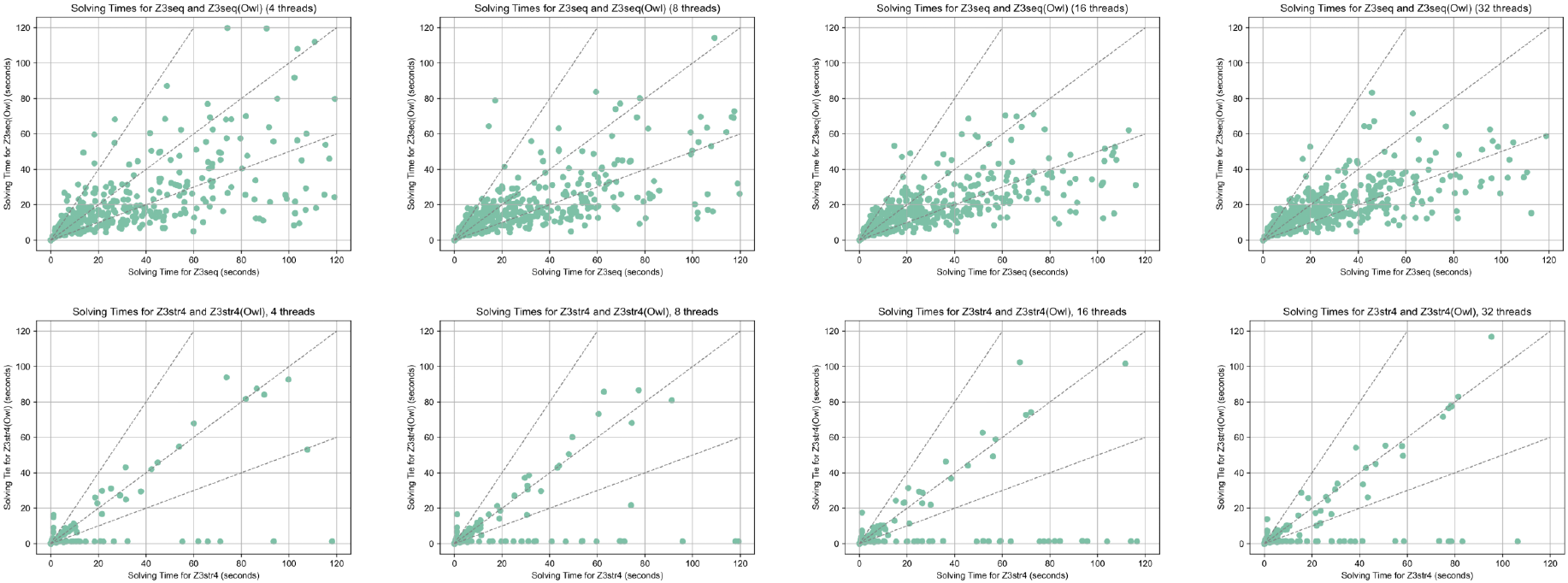}
	\caption{Scatter plot for Z3seq, Z3seq(\ToolName), Z3str4 and Z3str4(\ToolName)}
	\label{fig:scatter}
\end{figure*}

\smallskip
\noindent \textbf{Increased Numbers of Solved Instances}. Overall, Owl consistently improves solving capacity across all thread configurations. With four threads, Z3seq(Owl) solves 46 additional formulas (4.8\% increase), while Z3str4(Owl) solves nine more formulas compared to their baseline versions. These improvements are sustained across different thread counts, with Z3seq(Owl) maintaining a 1.4-2.6\% advantage with 32 threads.

\smallskip
\noindent \textbf{Reduced Runtime}.
On average, Z3seq(\ToolName) achieves a 1.44$\times$ speedup over Z3seq. The improvement is most pronounced in configurations with 16 and 32 threads, where parallelism is more effectively exploited. Figure~\ref{fig:scatter} illustrates the distribution of runtime reductions: for Z3seq(\ToolName), a substantial number of benchmarks are solved in less than half the time required by the baseline, with the scatter plot skewed toward faster solves. Z3str4(\ToolName) exhibits a different pattern, with notable gains on a smaller set of particularly hard instances, which it solves in a fraction of the time.

\smallskip
\noindent \textbf{Diminishing Returns with More Threads}.
The marginal benefit decreases with higher thread counts. For instance, the performance gain from 8 to 16 threads is smaller than that from 4 to 8. This trend likely stems from limitations in the parallel solver infrastructure, including load imbalance and the overhead of coordinating multiple threads. Additionally, the sequential solver's efficiency bounds the overall parallel speedup.

\begin{figure}
    \centering
    \begin{minipage}{0.495\textwidth}
        \centering
        \includegraphics[width=\textwidth]{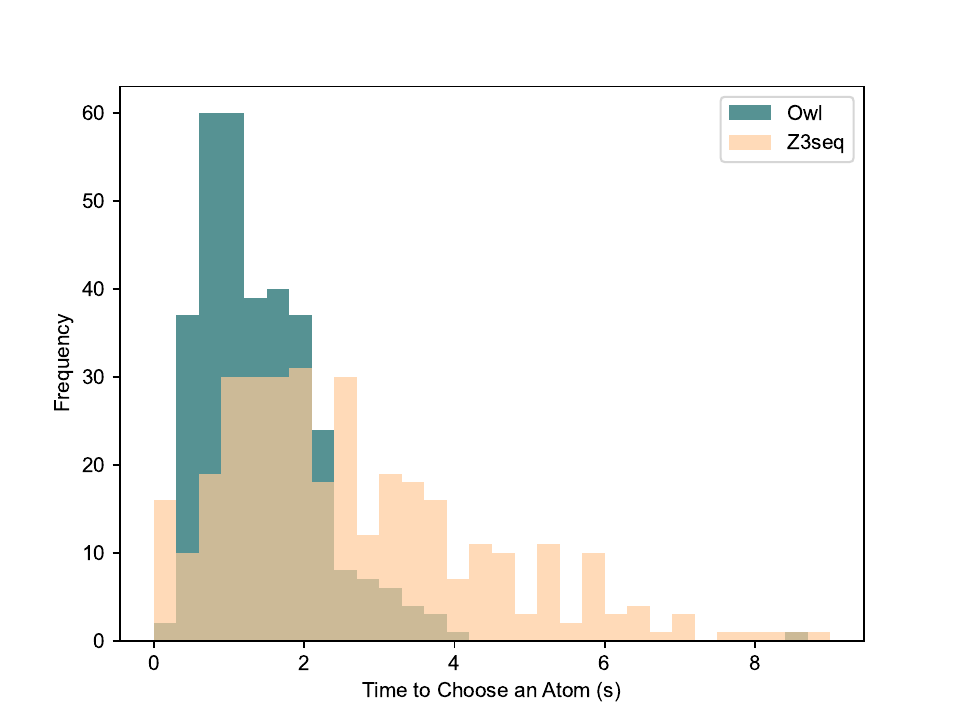}
        \caption{The overhead distribution to choose an atom in Owl and the default Z3 solver.}
        \label{fig:overhead}
    \end{minipage}
    \hfill
    \begin{minipage}{0.495\textwidth}
        \centering
        \includegraphics[width=\textwidth]{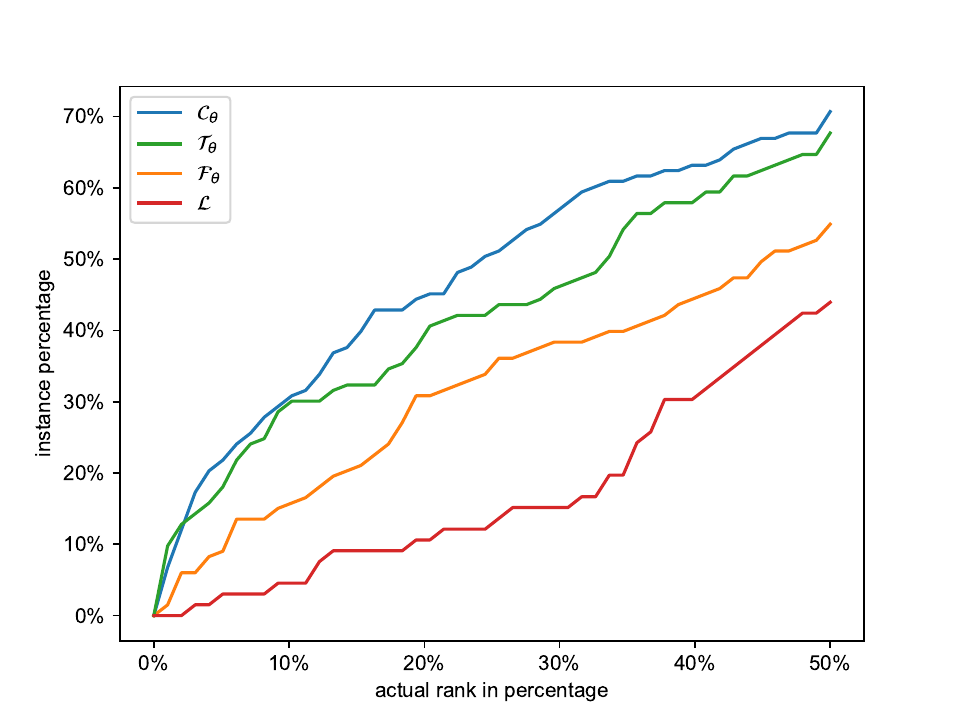}
        \caption{The accumulated percentage of instances against the predicted winner's actual rank (as a percentage).}
        \label{fig:rank-percentage}
    \end{minipage}
\end{figure}

\mybox{\textbf{Answer to RQ1}: \ToolName\ significantly improves Z3seq and Z3str4 under parallel cube-and-conquer, both in the number of solves and average runtime.}

\subsection{Performance of the Learned Models (RQ2)}
\label{subsec:eval-performance}
We analyze the models' effectiveness at predicting the atoms and their overhead.

\smallskip
\noindent \textbf{Atom Selection Accuracy}.
First, we evaluate the effectiveness of the learned models in selecting the optimal atom. 
To assess the ability of the learned models to rank splitting atoms, we compare predicted rankings with actual rankings based on solution times. We evaluate the learned models considered in RQ3, together with an additional method, $\mathcal{L}$, that employs the default lookahead heuristic in Z3.

Figure~\ref{fig:rank-percentage} illustrates the percentage of formulas where the predicted ``best'' splitting atom ranks within the top-$n\%$ of actual atoms. 
\begin{itemize}
    \item For Z3seq, the vanilla lookahead heuristic performs no better than a random selection. However, the regression-based models, particularly \ToolName, outperform all other methods, achieving a prediction accuracy of nearly 45\% for the top 20\% atoms.
    \item For Z3str4, a similar trend is observed, although the model’s performance is slightly diminished due to the solver’s limited handling of certain atoms. Nevertheless, the regression-based models still outperform the lookahead heuristic, achieving around 95\% accuracy in predicting the top 30\% of atoms.
\end{itemize}

The results reveal the general ordering of the predicted atoms: better choices appear closer to the front of the expected rank. And the top atoms in our predicted ranking overlap considerably with those in the actual ranking.

\smallskip
\noindent \textbf{Prediction Overhead}.
Figure~\ref{fig:overhead} provides an in-depth breakdown of the overhead distribution, illustrating that the additional computation required by \ToolName\ typically remains below 2 seconds and is generally lower than the overhead observed for the default lookahead heuristic. 
The overhead introduced by calculating feature values and executing the learned models accounts for only 23.6\% of the total solving time, slightly higher than the 21.4\% for the default lookahead heuristic. However, this difference is negligible compared to the performance gains.

\smallskip
\noindent \textbf{Feature Importance}.
We evaluate how various problem features influence the effectiveness of our learned models across different solvers. By identifying the most important features, we can better understand the factors contributing to solver efficiency and more effectively target optimization efforts.
We rank the problem features by their importance using random-forest feature importance~\cite{liaw2002classification,breiman2001random}, which aggregates the reduction in prediction error contributed by splits on each feature across the ensemble. 

\begin{table}[t]
\centering
\caption{Top-10 features for Z3seq(\ToolName) and Z3str4(\ToolName).}
\label{tbl:topfeat}
   \resizebox{0.49\textwidth}{!}
   {
\begin{tabular}{c c c}
\toprule
\textbf{Rank} & \textbf{Z3seq(Owl)} & \textbf{Z3str4(Owl)} \\ 
\midrule
1    & average variable degree & \# of conflict / \# of decision   \\ 
2    & \# of sub atoms in $p$  & \# of atoms in the formula   \\ 
3    & \# of $p$ in the formula   & lookahead score from Z3   \\ 
4    & \# of re.union in $p$  & \# of str.increment in $p$   \\ 
5    & \# of str.increment in $p$  & \# of sub atoms in $p$   \\ 
6    & \# of str.to\_re in $p$  & \# of assignments    \\ 
7    & \# of assignments  & \# of Int in $p$   \\ 
8    & \# of re.* in $p$  & \# of str.to\_re in $p$   \\ 
9    & lookahead score from Z3  & \# of re.union in $p$ \\ 
10   & \# of Char in $p$  & average variable degree   \\ 
\bottomrule
\end{tabular}
  }
\end{table}

Table~\ref{tbl:topfeat} presents the ten most important features identified for both the Z3seq(\ToolName) and Z3str4(\ToolName) models.
We observe several key insights:
\begin{itemize}
    \item  First, the ranking of the top 10 features differs significantly between solvers. Among the top ten features, seven are common to Z3seq and Z3str4, while the others vary in inclusion and order. This discrepancy suggests that each solver employs distinct heuristics and problem characteristics to optimize its performance. Consequently, solver-specific models are necessary to exploit each solver's most relevant features fully.
   \item Second, despite the variability in feature importance, certain features demonstrate consistent significance across both solvers. 
    These include the overall \# of assignments, the lookahead score from Z3, and the \# of \texttt{str.increment} occurrences in $p$. Such features align with the fundamental principles of lookahead and lookback splitting heuristics.
    Their consistent importance suggests that these aspects of the problem space play a critical role in solver performance, regardless of the underlying theory solver.
\end{itemize}

\mybox{\textbf{Answer to RQ2}:  Our learned regression models are highly effective at selecting splitting
atoms that rank near the top in solving time. While features influencing performance vary
across solvers, some critical features remain consistently important, offering valuable insights for further optimizations.}

\subsection{Ablation Study (RQ3)}
\label{subsec:eval-ml}

Here, we evaluate the performance \ToolName\ against two alternative methods for predicting splitting atoms. 

\begin{figure}[t]
    \centering
    \includegraphics[width=0.48\textwidth]{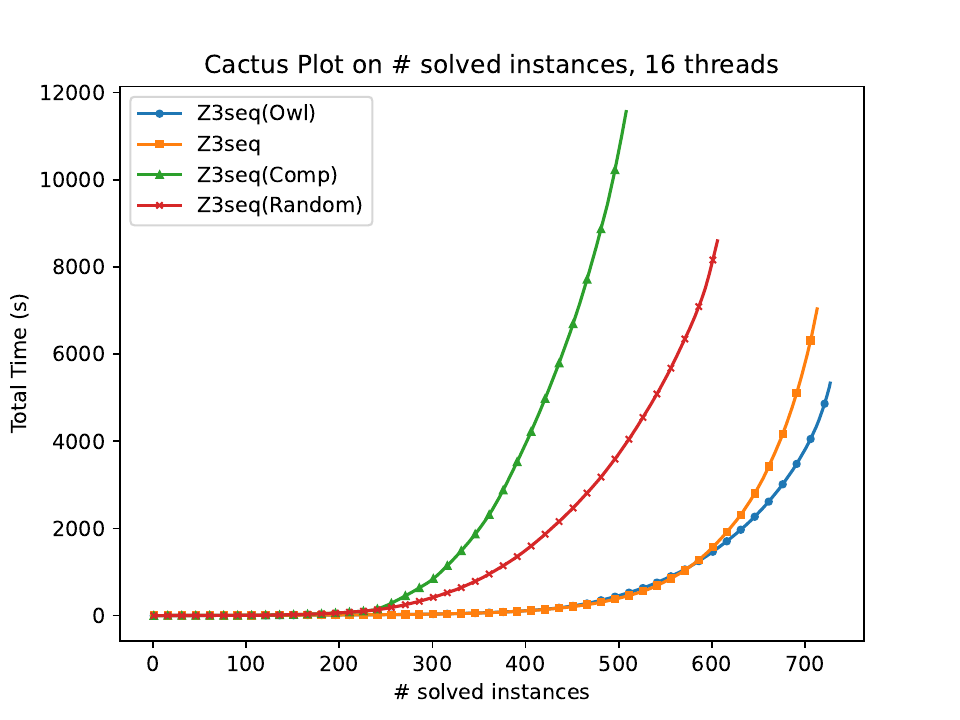}
    \caption{Cactus plot for Z3seq, Z3seq(Owl), Z3seq(Comparator) and Z3seq(Random)}
    \label{fig:ablation}
\end{figure}

\begin{figure*}[t]
	\centering
	\subfigure[Precision comparison.\label{fig:prec-comparison}]%
		{\includegraphics[width=0.48\textwidth]{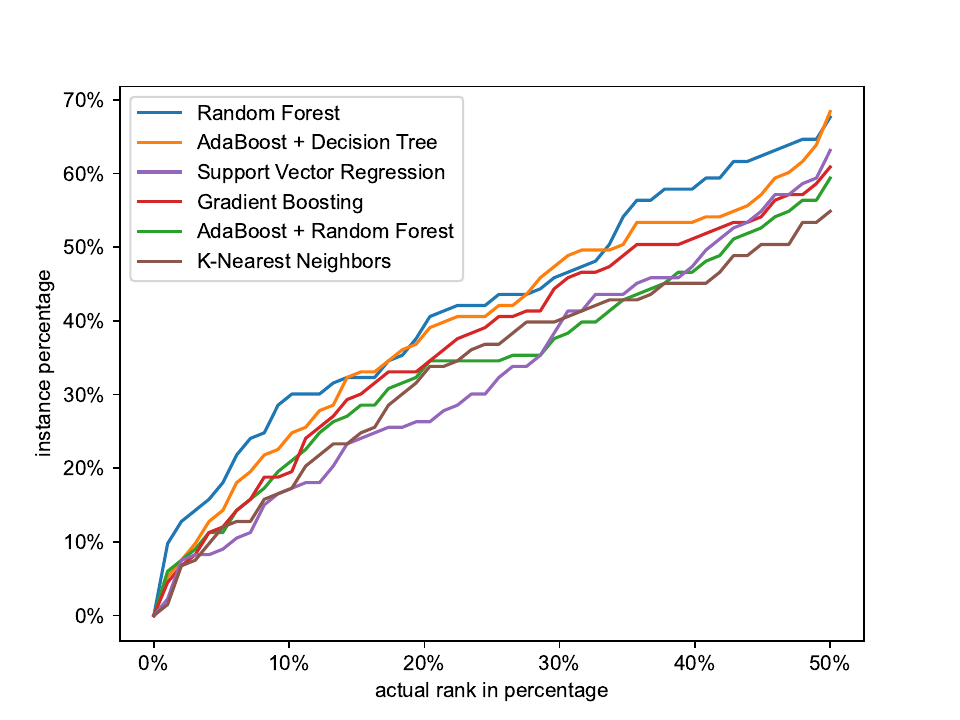}%
	}
	\hfill
	\subfigure[Time comparison.\label{fig:time-comparison}]%
		{\includegraphics[width=0.48\textwidth]{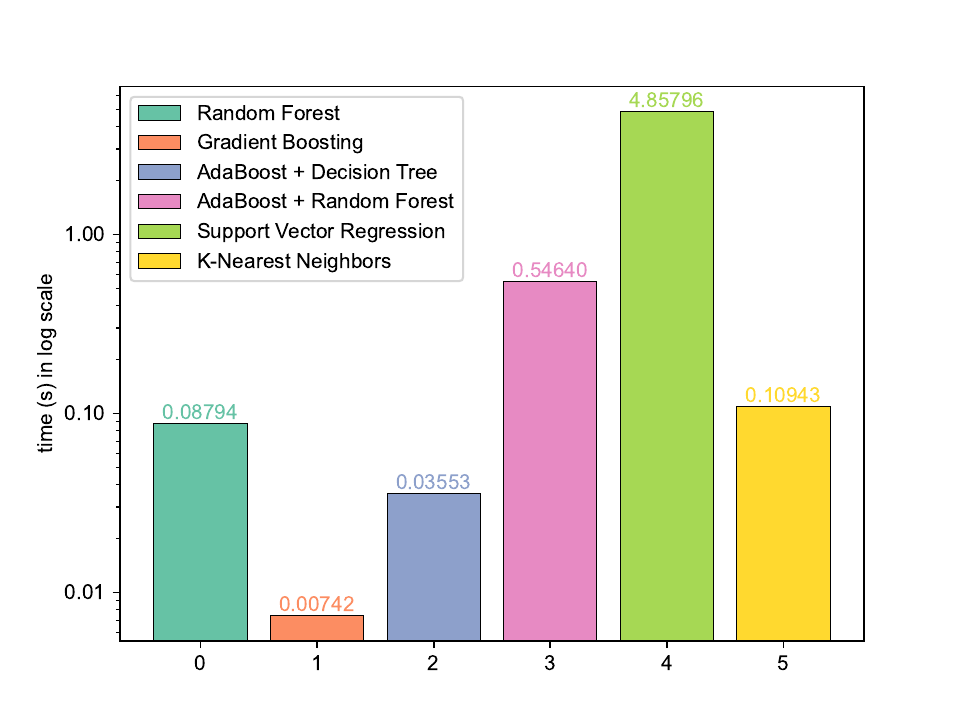}%
	}
	\caption{Comparison of different machine learning models}
	\label{fig:model-selection}
\end{figure*}

\smallskip
\noindent \textbf{Comparison of \ToolName\ and \ToolName-Comparator}.
First, we compare a variant that uses a binary classifier for pairwise atom comparisons, as in prior SAT-solving research (\cref{subsec:existing}).
We present the results of this comparison in Figures~\ref{fig:model-selection} and~\ref{fig:ablation}, and highlight several key observations.

\begin{itemize}
    \item \emph{End-to-end Improvements}: The pairwise-classification baseline does not translate into the same end-to-end gains as \ToolName. Figure~\ref{fig:ablation} shows that \ToolName\ reduces the overall solving time by 29\%, while Comparator and Random both perform worse than Z3seq.
    \item \emph{Optimal Atom Prediction}:  
    The regression model employed by \ToolName\ more reliably identifies promising atoms for splitting, thereby improving solver performance. Figure~\ref{fig:model-selection} shows that the random-forest regressor used in \ToolName\ provides a better accuracy-efficiency trade-off than the pairwise comparator models.
    \item  \emph{Additional Runtime Overhead}:
 The Comparator model incurs significant overhead during the linear scan, indicating that \ToolName\ strikes a better balance between performance gains and computational efficiency.
\end{itemize}

\noindent \textbf{Variants of \ToolName\ with different Regressor}.
The choice of learning model may influence the performance of both \ToolName\ and Comparator. For \ToolName, we compare several regressors; for Comparator, we compare several binary classifiers. Figure~\ref{fig:model-selection}(a) reports prediction quality, and Figure~\ref{fig:model-selection}(b) reports the corresponding inference time. Overall, random-forest regression provides the best trade-off for \ToolName: some alternative models offer comparable prediction quality, but none yield a better end-to-end balance of ranking quality and runtime overhead. For Comparator, changing the classifier has only a limited effect on overall solving time because the pairwise-comparison design still requires repeated model invocations during atom ranking.

\mybox{\textbf{Answer to RQ3}: The random forest regressor demonstrates superior accuracy and efficiency in selecting atoms.}

\section{Related Work}
\label{sec:related}

 \noindent  \textbf{String Constraint Solving}.
There is a vast amount of literature on string constraint solving, which can be broadly categorized into bounded and unbounded methods. 
The bounded methods often assume
that the string variables have fixed lengths, such as
HAMPI~\cite{kiezun2009hampi}, CFGAnalyzer~\cite{axelsson2008analyzing}, and \cite{he2013solving}.
However, strings in many programming languages are variable-length, necessitating reasoning about unbounded strings.
For example, \cite{saxena2010symbolic} show that there is still a big gap in applying bounded methods to constraints arising from the analysis of Web applications because the constraints usually involve unbounded strings.
In comparison, the unbounded methods are often based on 
the theory of automata or regular language,
such as 
ABC~\cite{aydin2015automata}, Stranger~\cite{yu2010stranger}, Norn~\cite{abdulla2015norn},
Sloth~\cite{holik2017string}, and Ostrich~\cite{chen2019decision}.
The most recent solvers for unbounded strings are built on the CDCL($T$) framework, which enables combining theories such as strings and integers. Examples include CVC4~\cite{barrett2011cvc4,liang2014dpll,reynolds2017scaling,reynolds2018rewrites,reynolds2019high}, S3~\cite{trinh2014s3}, Z3seq~\cite{z3seq}, Trau~\cite{aziz2017flatten,abdulla2019chain}, Z3str~\cite{zheng2013z3}, Z3str2~\cite{zheng2017z3str2}, Z3str3~\cite{Z3Str33}, and others.
\cite{thome2017search} solve string constraints via ant colony optimization, which complements CVC4 and Z3 in modeling some special string APIs in Java Web applications.
Our approach builds on CDCL($T$) and uses the sequential solver's algorithmic components.
A side benefit is that \ToolName\ can handle all the string operations supported by the underlying solvers, e.g., Z3seq and Z3str4.

\smallskip
\noindent  \textbf{Parallel Constraint Solving}.
Generally, two main parallel-solving approaches have been developed: the portfolio and divide-and-conquer approaches.
The portfolio approach
runs several solvers in parallel on the same input formula 
and obtains the solution from the first solver that succeeds~\cite{xu2008satzilla,DBLP:conf/cav/WintersteigerHM09}.
Employing a portfolio of SMT solvers has been explored in~\cite{palikareva2013multi} for KLEE, in which each solver runs independently, and the result of the fastest is taken. 
The divide-and-conquer approach, on the other hand,
splits the search space of the input formula into different parts, which are solved in parallel~\cite{DBLP:conf/sat/HyvarinenMS15,paropensmt,DBLP:conf/atva/MarescottiHS16,cheng2018parallelizing}.
A crucial algorithmic component of divide-and-conquer solvers is the splitting heuristic for the search space.
In the existing parallel SAT solving literature,
the heuristics can be broadly categorized as \emph{lookback-based}~\cite{audemard2016adaptive,le2019modular,nejati2017propagation} and \emph{lookahead-based}~\cite{heule2018cube,biere2017cadical}  approaches.
Lookback-based heuristics~\cite{audemard2016adaptive,le2019modular,nejati2017propagation} compute statistics on ``how well an atom participated in the search exploration in the past'', and rank them appropriately.
Lookahead-based heuristics~\cite{heule2018cube,biere2017cadical} analyze the impact of splitting on different atoms and rank the atoms based on information gathered during the propagation.
The most recent work on divide-and-conquer SMT solving is based on lookahead~\cite{DBLP:conf/sat/HyvarinenMS15,paropensmt,DBLP:conf/atva/MarescottiHS16,reisenberger2014pboolector}, utilizing hand-crafted splitting heuristics, and has not been applied to string constraints.
We aim to automatically learn the heuristics and 
use lookback and lookahead information as the dynamic feature.

\smallskip
\noindent \textbf{Data-Driven Constraint Solving}.
Data-driven techniques have been employed in various ways to accelerate constraint solvers.
First, the most common approach  is algorithm selection
~\cite{xu2008satzilla,bridge2011case,amadini2014sunny,hurley2014proteus,kadioglu2010isac,scott2020machsmt},
which aims to predict the best solver or solver configuration for a given formula.
Second, recent efforts have utilized Z3's tactic language to formulate tactic optimization as a program synthesis problem.
FastSMT~\cite{balunovic2018learning}  synthesizes a solving strategy as a loop-free program with branches.
\cite{chen2021synthesize} synthesize tailored solving strategies for symbolic execution through a two-stage process involving offline-trained models and online tuning.
\cite{lu2024layered} propose a Monte Carlo Tree Search-based method for synthesizing effective strategies.
Finally, search-based techniques have been used to solve constraints directly, as opposed to guiding the existing solving algorithms, such as particle-swarm optimization~\cite{souza2011coral}, gradient-based search~\cite{shen2019neuro,borzacchiello2021fuzzing}, ant colony optimization~\cite{thome2017search}, random walk~\cite{dinges2014solving}, evolutionary search~\cite{liew2019just}, Monte Carlo Markov Chain (MCMC)~\cite{fu2016xsat}, fuzzing~\cite{borzacchiello2021fuzzing,liew2019just}, and classification-based optimizations~\cite{li2016symbolic}.
Compared to prior work, we address a different problem: automatically selecting the splitting atom to speed up parallel string solvers.
The pair-wise voting strategy was previously used to rank a set of solvers in algorithm selection~\cite{xu2008satzilla}.

\section{Conclusion and Future Work}
\label{sec:conclusion}

We presented \ToolName, a data-driven approach to accelerating parallel string constraint solving through learned splitting heuristics. We integrated \ToolName\ into two state-of-the-art solvers, Z3seq and Z3str4, and demonstrated substantial performance improvements over existing techniques. A key design choice is framing atom selection as a regression task rather than pairwise classification, which preserves quantitative performance information and reduces cascading errors. As future work, we plan to explore embedding-based representations for richer feature modeling and to extend the learning framework to additional algorithmic components, such as scheduling policies.

We identify several directions for future work.
First, we plan to investigate embedding-based representations for feature modeling. Embeddings can capture formula structure and atom context more compactly than hand-crafted numeric features, potentially improving both generalization and scalability.
Second, we plan to extend the approach to additional solvers. We have already integrated our technique with Z3str3~\cite{Z3Str33}; however, Z3str3 lacks support for several string operations present in our benchmarks, and we encountered multiple bugs during integration. We also considered CVC5, but it does not currently provide a cube-and-conquer engine.
Third, we aim to apply the same regression-based splitting strategy to other cube-and-conquer settings—for example, parallel SMT solving for linear arithmetic—and to study whether learned heuristics transfer across theories and benchmark suites

\bibliographystyle{ACM-Reference-Format}
\bibliography{sigproc}
\end{document}